\documentclass{article}

\usepackage{arxiv}

\usepackage{comment}

\usepackage[utf8]{inputenc} 
\usepackage[T1]{fontenc}    
\usepackage{hyperref}       
\usepackage{url}            
\usepackage{booktabs}       
\usepackage{amsfonts}       
\usepackage{nicefrac}       
\usepackage{microtype}      
\usepackage{lipsum}
\usepackage{float} 
\usepackage{graphicx}
\usepackage{amsmath}
\graphicspath{ {./images/} }

\usepackage[numbers]{natbib}   
\usepackage{hyperref}

\title{Evaluating a Multi-Agent Voice-Enabled Smart Speaker for Care Homes: A Safety-Focused Framework}

\author{
  Zeinab Dehghani \\
  University of Hull, UK \\
  \texttt{z.dehghani-2023@hull.ac.uk} \\
  \And
  Rameez Raja Kureshi \\
  University of Hull, UK \\
  \texttt{r.kureshi@hull.ac.uk} \\
  \And
  Koorosh Aslansefat \\
  University of Hull, UK \\
  \texttt{k.aslansefat@hull.ac.uk} \\
  \And
  Faezeh Alsadat Abedi  \\
  University of Southampton, UK \\
  \texttt{f.a.abedi@soton.ac.uk} \\
  \And
  Dhavalkumar Thakker \\
  University of Hull, UK \\
  \texttt{d.thakker@hull.ac.uk} \\
  \And	
  Lisa Greaves\\
  Connexin, Hull, UK \\
  \texttt{ l.greaves@connexin.co.uk} \\
  \And	
  Bhupesh Kumar Mishra\\
  University of Hull, UK \\
  \texttt{ bhupesh.mishra@hull.ac.uk} \\
  \And	
  Baseer Ahmad\\
  University of Hull, UK \\
  \texttt{ baseer.ahmad@hull.ac.uk} \\
  \And	
  Tanaya Maslekar\\
  Leeds Teaching Hospital NHS Trust, UK \\
  \texttt{ tanaya.maslekar4@nhs.net} \\
}

\begin{document}
\maketitle
\begin{abstract}

Artificial intelligence (AI) is increasingly being explored in health and social care to reduce administrative workload and allow staff to spend more time on patient care. This paper evaluates a voice-enabled Care Home Smart Speaker designed to support everyday activities in residential care homes, including spoken access to resident records, reminders, and scheduling tasks. A safety-focused evaluation framework is presented that examines the system end-to-end, combining Whisper-based speech recognition with retrieval-augmented generation (RAG) approaches (hybrid, sparse, and dense). Using supervised care-home trials and controlled testing, we evaluated 330 spoken transcripts across 11 care categories, including 184 reminder-containing interactions. These evaluations focus on (i) correct identification of residents and care categories, (ii) reminder recognition and extraction, and (iii) end-to-end scheduling correctness under uncertainty (including safe deferral/clarification). Given the safety-critical nature of care homes, particular attention is also paid to reliability in noisy environments and across diverse accents, supported by confidence scoring, clarification prompts, and human-in-the-loop oversight. In the best-performing configuration (GPT-5.2), resident ID and care category matching reached 100\% (95\% CI: 98.86--100), while reminder recognition reached 89.09\% (95\% CI: 83.81--92.80) with zero missed reminders (100\% recall) but some false positives. End-to-end scheduling via calendar integration achieved 84.65\% exact reminder-count agreement (95\% CI: 78.00--89.56), indicating remaining edge cases in converting informal spoken instructions into actionable events. The findings suggest that voice-enabled systems, when carefully evaluated and appropriately safeguarded, can support accurate documentation, effective task management, and trustworthy use of AI in care home settings.
\end{abstract}


\section{Introduction}

In residential care homes, staff routinely balance direct care with substantial administrative work, including recording daily observations, documenting care activities, and managing reminders for tasks such as medication, mobility support, and personal care. These activities are often carried out under time restrictions, in busy and noisy environments, and alongside competing demands on staff attention. As a result, documentation may be delayed, incomplete, or completed retrospectively, increasing the risk of missed information and reducing the time available for meaningful interaction with residents~\cite{bjerkan2021patient}. This brings the need for administrative assistant tools in residential care homes to support the care workers' day-to-day work. In recent years, AI has been increasingly explored in health and social care to reduce this administrative burden and support care staff in managing routine yet safety-critical tasks~\cite{edwards2021use,merkel2025identification,topol2019deep}. Among these developments, voice-based technologies such as smart speakers have attracted particular interest, as they allow hands-free interaction and can be used while staff are actively engaged in care. Studies have suggested that smart speakers can support care documentation, improve access to information, provide timely reminders, and enhance engagement for both staff and residents~\cite{laranjo2018conversational, saripalle2024command}. One example of this approach is the Speech-Controlled Smart Speaker for Accurate, Real-Time Health and Care Record Management~\cite{carrick2025speech}, which introduced a purpose-built smart speaker to support voice-based interaction with care records. That work focused specifically on strengthening the foundation of automatic speech recognition (ASR) required for reliable use in care homes, where background noise and accent diversity pose significant challenges. By fine-tuning the Whisper model on British regional accents and introducing real-time safeguards to prevent speech hallucinations, this study demonstrated substantial reductions in word error rates and improved robustness in realistic care-like environments. This work established the technical reliability of speech input as a prerequisite for safe voice-enabled care systems, providing the foundation for more complex, task-oriented functionality. 

Despite growing interest, evidence remains limited on whether voice-enabled systems can deliver end-to-end reliability (speech → structured record → retrieval → action) under realistic care-home conditions, where errors can propagate across stages. Besides, deploying voice-enabled systems in care homes introduces distinct risks as care homes are safety-critical environments, where failures in communication or reliability may lead to delayed interventions, missed care tasks, medication errors, or serious harm. Errors such as mistranscribed reminders, missed notifications, or incorrect information retrieval can propagate through the system and have disproportionate consequences for vulnerable residents. A further challenge arises from linguistic diversity since care staff and residents may speak with varied accents, dialects, or speech impairments, and even minor transcription errors can result in misreported symptoms or incorrect care instructions. For this reason, rigorous and multi-level evaluation is essential before real-world deployment. Furthermore, beyond technical considerations, broader governance and deployment challenges must also be acknowledged in the care home administrative work. Recent NHS England guidance on AI-enabled ambient scribing systems highlights risks including output inaccuracies, system downtime, and poor integration with existing digital record infrastructures~\cite{sarvari2025challenges}. On top of these challenges, Generative AI components may also unintentionally exceed their intended role, for example, by producing unsanctioned suggestions, raising concerns around regulation, accountability, and liability~\cite{okonji2024applications}. Moreover, the AI-assistance in care home administrative work comes with other additional issues as well, including data privacy and security, accent-related bias, workflow disruption due to limited interoperability, and uncertainty around responsibility in safety-critical use cases~\cite{ehrlich2025forestgpt}. 

Taken together, these concerns reinforce the need for transparent, evidence-based evaluation frameworks to support the responsible adoption of voice-enabled AI in health and social care. Building on the Speech-Controlled Smart Speaker for Accurate, Real-Time Health and Care Record Management foundation, the present study extends the Care Home Smart Speaker into a multi-agent system designed to support everyday care workflows. The architecture comprises specialised components for speech transcription (Whisper-based ASR), natural language parsing, structured data storage (PostgreSQL), retrieval-based question answering using sparse, dense, and hybrid methods, reminder scheduling, Google Calendar integration, and smart speaker notifications with follow-up confirmations. This study integrates Whisper-based ASR into the smart speaker framework, aiming to achieve robust transcription performance across diverse voices and noisy care environments. Throughout, the system is treated as a safety-critical pipeline in which upstream uncertainty (e.g., transcription ambiguity) must be detected and controlled to avoid downstream unsafe actions. This paper also proposes a safety-focused evaluation framework that assesses the Care Home Smart Speaker across multiple dimensions, including (i) structured data parsing and integrity, (ii) reliable retrieval of resident and care-category information, (iii) reminder extraction and scheduling behaviour, and (iv) uncertainty-handling safeguards (confidence scoring, clarification prompts, and human-in-the-loop confirmation). Unlike evaluations that focus on isolated component metrics ~\cite{carrick2025speech}, the framework is explicitly designed to detect error propagation and to treat “safe deferral/clarification” as a valid outcome when inputs are ambiguous. To ground evaluation in realistic usage, the study analyses 330 logged spoken interactions across 11 care categories, including 184 reminder-containing interactions, collected from supervised care-home trials and controlled testing. Performance is reported using proportion-based correctness metrics with Wilson confidence intervals, along with semantic distance measures to assess meaning preservation, where relevant (defined in Section 5). By combining accuracy-based measures (distance-based metrics such as cosine similarity, Wasserstein distance) with semantic similarity metrics and statistical confidence intervals, the framework provides a structured and transparent approach to evaluating performance in safety-critical care settings~\cite{belisle2024stakeholder, pacyna2025patient}.

Beyond technical accuracy, this study additionally considers the system’s ability to manage uncertainty, support explanation of outputs, and foster trustworthy interaction between staff and technology. These qualities are particularly important in care environments, where errors may directly affect vulnerable individuals~\cite{mittelstadt2019principles}. By integrating retrieval-focused and task-oriented evaluation, this work contributes not only to system-level assessment but also to wider discussions around the responsible use of AI in health and social care. In summary, this study makes four key contributions:
\begin{enumerate}
\item It presents a safety-focused evaluation framework for multi-agent smart speaker systems in care homes, extending prior work on speech-controlled care technologies by providing a single, integrated approach to evaluating the reliability of transcription, parsing, retrieval, and scheduling.
\item It introduces evaluation to assess the retrieval of rare but safety-critical information from care records, addressing a failure mode that is particularly relevant to retrieval-augmented systems but rarely examined in care settings.
\item It demonstrates the importance of robust speech recognition across diverse accents, dialects, and noisy environments, directly responding to concerns about bias, exclusion, and inequitable performance in voice-based care technologies.
\item It advances responsible AI practice in health and social care by illustrating how confidence scoring, clarification prompts, and human-in-the-loop verification can be used to reduce risks associated with output inaccuracies, misuse of systems, and uncertainty around accountability.
\end{enumerate}

The rest of  Section 2 reviews related work and identifies gaps in end-to-end evaluation for safety-critical voice systems. Section 3 describes the system architecture and safety design choices. Sections 4 and 5 define the problem definition and evaluation objectives, assurance argument, and metrics. Section 6 describes the evaluation dimensions. Section 7 presents the experimental setup and result analysis, followed by discussion and limitations in Section 8 and conclusion and future work in Section 9.

\section{Literature Review}

The rapid growth of voice-activated smart speakers, such as Amazon Echo and Google Home, has generated increasing interest in their potential applications in healthcare, elderly care, and disability support. Over the past decade, these devices have increasingly been considered as tools to support independence, communication, and well-being in care contexts. This interest also reflects broader pressures within health and social care systems, where staff time is limited, and documentation demands are high, creating a need for practical technologies that can fit into everyday workflows rather than adding further burden. However, the evidence base remains uneven. Many studies focus on consumer-oriented use cases, such as companionship, entertainment, and simple reminders, whereas relatively few examine end-to-end reliability for safety-critical workflows such as care documentation, record retrieval, and task scheduling. Two broad themes emerge in the literature: (i) how smart speakers are used in care settings, and (ii) how their effectiveness, acceptability, and limitations have been evaluated. More recently, a further theme has emerged around LLM-enabled voice assistants and the corresponding need for evaluation that extends beyond usability and transcription accuracy.

\subsection{Use of Smart Speakers in Care}

Several recent reviews have mapped the role of smart speakers in health and social care. Saripalle and Patel~\cite{saripalle2024} conducted a scoping review of 59 studies and identified applications ranging from independent living support and remote monitoring to loneliness reduction and improved access to medical information for providers. Merkel and Schorr~\cite{merkel2025} similarly reported that smart speakers are being adopted across home care, hospitals, and long-term care, with older adults and patients as the primary users and clinicians as a secondary user group. In addition, empirical studies provide further insight into these applications. Quinn et al.~\cite{quinn2024assessing} tested an Alexa-based intervention to promote physical activity among older adults and found that participants rated the technology as highly usable and expressed strong intentions to continue using it, although some required support with connectivity and command phrasing. Astell and Clayton~\cite{astell2024like} showed that smart speakers reduced feelings of loneliness among very old adults in supported housing, with participants describing the devices as both a comforting presence and a source of control over daily routines. These community and long-term care case studies offer similar suggestions: perceived benefits often depend on staff support, device training, and the ability to adapt interactions to residents’ needs over time, rather than on simple “plug-and-play” deployment.

Research involving people with intellectual disabilities has also reported positive outcomes. Smith et al.~\cite{smith2023smart} found that participants felt smart speakers increased their agency and independence, despite difficulties with pronunciation and remembering commands. This is consistent with broader human--computer interaction research describing voice assistants as “accidental accessibility tools,” offering independence-supporting functions even when they were not originally designed for disability use~\cite{pradhan2018accessibility}. In hospital settings, Franco et al.~\cite{franco2023exploratory} found that smart speakers were actively used in emergency departments during the COVID-19 pandemic, supporting both clinical communication and patient entertainment. Edwards et al.~\cite{edwards2021use} reported that the introduction of smart speakers in UK care homes was feasible and well received, providing further evidence of their potential to enhance daily living in residential care environments. Their regional implementation study also highlighted how devices were used in practice, the barriers to adoption, and the importance of organisational readiness and staff engagement.

Patterns of adoption are not uniform. Nimrod and Edan~\cite{nimrod2022technology}, in a longitudinal study of older women, identified three distinct patterns of domestication: broad domestication, in which the device became integrated into daily life; focused domestication, in which use centred on one or two functions; and restrained domestication, in which use became occasional after an initial trial period. Other long-term usage studies similarly suggest that engagement often stabilises around a small number of core functions, such as timers, music, weather, and reminders, with implications for the sustainability of health- and care-related applications. Han et al.~\cite{chen2023older} further highlighted that onboarding and setup remain significant barriers for older users, many of whom struggled to connect devices to Wi-Fi or complete installation without assistance. The recent work has explored advanced integrations between voice interfaces and AI systems. Carrick et al.~\cite{carrick2025speech} proposed a speech-controlled smart speaker system for the accurate, real-time management of health and care records, while Yang et al.~\cite{yang2024talk2care} introduced Talk2Care, a Large Language Model (LLM)-based voice assistant designed to improve communication between healthcare providers and older adults. Similarly, Lima et al.~\cite{lima2025promoting} investigated socially assistive robots powered by LLMs to promote cognitive health in elder care. These studies point to a growing convergence between voice interfaces and AI-driven healthcare technologies. Treder et al.~\cite{treder2024introduction} further discuss the role of LLMs in dementia care and research, highlighting their potential to augment future smart speaker-based interventions. At the same time, this shift toward LLM-enabled systems increases the importance of careful evaluation, since errors may become more difficult to detect and may occur downstream, for example, during retrieval or task execution, rather than only at the transcription stage.

\subsection{Evaluation of Smart Speaker Systems in Care Settings}

Studies evaluating smart speakers in care settings generally report high levels of interest and acceptability, but they also identify important limitations. Kim~\cite{kim2021exploring}, examining first-time interactions between older adults and Alexa, found that participants appreciated the convenience of voice commands but often experienced frustration with recognition errors, misunderstood the system’s capabilities, and expressed privacy concerns. Similarly, Smith et al.~\cite{smith2023smart} reported positive subjective experiences of empowerment and independence among people with intellectual disabilities, although standardised wellbeing measures showed little measurable change. Technical challenges are particularly salient for users with speech impairments. Ballati et al.~\cite{ballati2018hey} showed that commercial voice assistants performed poorly with dysarthric speech, achieving recognition accuracy of only around 50-60 percent and highlighting a significant accessibility gap. More recent speech and language research continues to report inconsistent ASR performance for people with speech and language disorders, reinforcing the need for inclusive evaluation rather than assuming that performance observed in general populations will transfer to care settings~\cite{hui2025enhancing, green2003automatic}.

Ethical concerns have also received substantial attention. Lau, Zimmerman, and Schaub~\cite{lau2018alexa} found that many users worry about surveillance and constant listening, with some adopting protective strategies such as muting microphones or unplugging devices. These concerns are amplified in healthcare settings, where sensitive personal and health-related information may be disclosed. Studies of privacy perceptions among older adults likewise highlight uncertainty about what voice assistants record, who can access the data, and how the data are used, all of which may affect willingness to adopt these tools~\cite{bonilla2020older}. Foundational work by Pradhan et al. ~\cite{pradhan2018accessibility} contextualises these concerns by showing that smart speakers often function as “accidental accessibility tools,” enabling people with disabilities to complete tasks more independently, even though the devices were not explicitly designed for this purpose. Bentley et al.~\cite{bentley2018understanding} examined long-term household use and found that engagement typically stabilised around a few key functions, such as playing music, checking the weather, and setting timers. In care contexts, this suggests that sustained adoption is likely to depend on whether health and social care functions can become part of routine practice rather than remaining short-lived novelty features. Beyond usability and adoption, some of the studies have also raised safety concerns about speech-to-text systems themselves. For example, Koenecke et al.~\cite{koenecke2020racial} reported that some state-of-the-art speech recognition systems can occasionally generate hallucinated phrases that were not present in the original audio, which is particularly concerning when transcripts are treated as formal records. This highlights the need for safeguards in care environments, including confidence scoring, verification, and clear limits on what systems are permitted to automate.

Reviewing the literature indicates that there is a growing industry interest in this area. Oakland Care and Vocala~\cite{oakland2024voicepilot} reported results from a voice technology trial in UK care homes, while Aston University and Lee Mount Healthcare~\cite{aston2025smartcarehome} announced the development of an AI-powered smart care home system intended to improve residential care quality. Pilot deployment reports from domiciliary and residential care settings~\cite{edwards2021use} similarly suggest that practical trials are already underway, although findings remain underrepresented in the academic literature. In health and social care, this gap between rapid real-world experimentation and slower academic reporting strengthens the case for transparent and reproducible evaluation approaches that can be shared and scrutinised. Taken together, the literature suggests that smart speakers have considerable potential to support care through increased independence, social connection, and practical assistance. However, important challenges remain, including accessibility barriers, uneven speech recognition performance for diverse and impaired voices, and persistent concerns about privacy and trust. As these systems evolve toward LLM-enabled, task-oriented assistants capable of retrieval and scheduling, evaluation must move beyond simple usability or transcription accuracy to address end-to-end reliability, error propagation, and rare but high-impact failure modes. This underscores the need for a rigorous, multi-dimensional evaluation framework, which is the central contribution of this study.

\section{System Overview and Architecture}

This study evaluates the Care Home Smart Speaker, a voice-enabled system designed to support care staff with spoken documentation, retrieval of care records, and task reminders through natural spoken interaction. The system is intended for use during routine care activities, allowing staff to record information or request updates without interrupting care delivery to interact with screens or keyboards. This section provides an overview of the system architecture and explains how information flows from spoken input to structured records and, where appropriate, to task execution.

\begin{figure}[ht]
  \centering
  \includegraphics[width=1\linewidth,keepaspectratio]{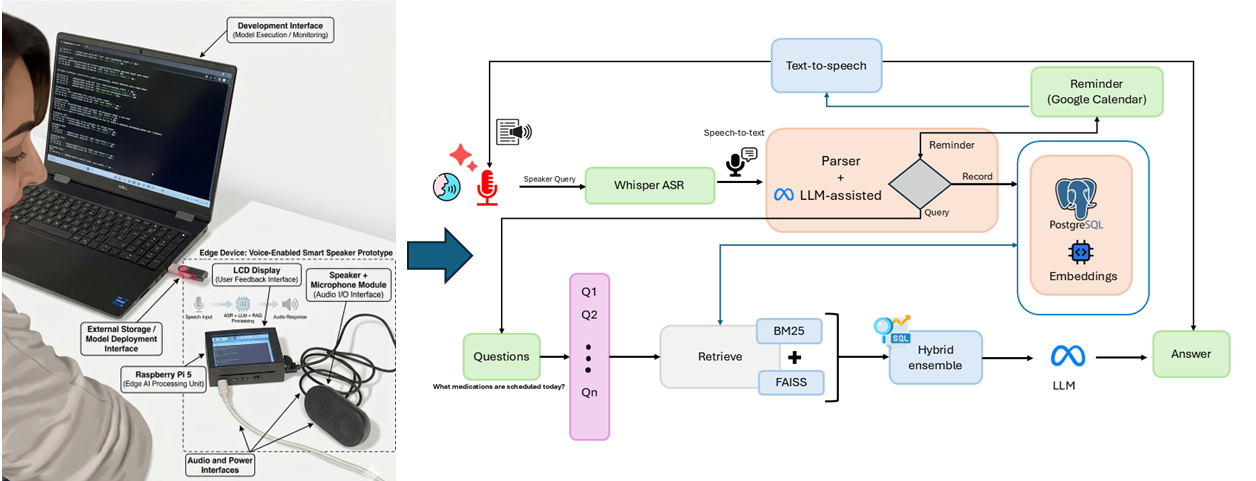}
  \caption{System overview and architecture of the voice-enabled care support platform.}
  \label{fig:rag}
\end{figure}
Figure~\ref{fig:rag}provides a high-level overview of the proposed voice-enabled care support platform and illustrates the end-to-end data flow from spoken user input to system response. The architecture adopts a modular and layered design, separating speech processing, intent understanding, structured storage, information retrieval, and response generation. This separation is intended to support transparency, auditability, and responsible use of AI within care and healthcare contexts.
\subsection{End-to-end workflow}
At a high level, the system operates as a pipeline that transforms spoken audio input into structured care records and, where appropriate, into scheduled tasks such as reminders. The workflow is deliberately designed as end-to-end, recognising that failures may occur not only during speech recognition but also during parsing, storage, retrieval, or task execution.

The main stages of the workflow are as follows.

\begin{enumerate}
\item \textbf{Spoken input and speech recognition:} Interaction begins when a member of care staff speaks to the smart speaker during routine care activities. Audio capture is initiated by an explicit interaction trigger (e.g., a wake word), rather than assuming unrestricted, continuous recording. Audio is continuously monitored and captured by the device and passed to an automatic speech recognition (ASR) component based on the Whisper model. Building on earlier work in care-home speech recognition, this component has been adapted to handle background noise and accent diversity commonly encountered in residential care settings. Real-time safeguards, including signal quality checks, are applied to reduce the risk of spurious or misleading transcriptions.

\item \textbf{Natural language parsing into a structured format:} The transcribed text is processed to extract structured information required for care documentation and task management. This includes identifying the intended Resident, the relevant care category, and any time-based instructions that may indicate a reminder or follow-up action.

\item \textbf{Structured storage:} Parsed content is stored in a structured database (PostgreSQL) using a predefined schema. Each entry includes a resident identifier, timestamp, care category, and free-text statement. Schema validation is applied before storage to prevent malformed or incomplete records from being inserted, supporting later retrieval, traceability, and auditing.

\item \textbf{Retrieval for spoken queries:} In addition to recording information, the system supports spoken queries from care staff, such as requests for recent observations or reminders associated with a particular resident. Relevant records are retrieved from the database using a retrieval component that supports sparse, dense, and hybrid strategies. This reflects the fact that care staff may phrase questions in different ways and that care records often contain a mix of structured and free-text information. Retrieved records are then passed to an LLM component, which synthesises the retrieved evidence and generates the final natural language response returned to the user through the smart speaker.

\item \textbf{Task handling and scheduling:} When spoken input includes a time-based instruction (e.g., ``Remind me to check blood pressure at 2 pm''), the system treats it as a reminder task. Task details are extracted during parsing and passed to a scheduling component, which validates timing information and determines whether the task can be scheduled immediately or requires clarification. Validated reminders are added to a calendar system that maintains an explicit audit trail of scheduled actions.

\item \textbf{Confirmation and feedback:} For both retrieval and task-related interactions, the system provides spoken feedback through the smart speaker. Where confidence is low or ambiguity is detected, clarification prompts are used, and actions may be deferred for human confirmation rather than executed automatically.
\end{enumerate}
\subsection{System components}
To make the system transparent for evaluation, a set of the following cooperating components, each responsible for a specific function, is considered.
\begin{itemize}
\item \textbf{ASR component (Whisper-based):} Converts speech to text and is designed to be robust to accent diversity and background noise. Analysing earlier systems developed in this domain demonstrated that care-relevant performance requires explicit attention to accent coverage and to preventing real-time hallucination.

\item \textbf{Parsing component (LLM-assisted):} Converts transcribed speech into a structured representation suitable for care records and reminders, including resident identification, care category assignment, and extraction of reminder attributes where present, along with schema validation that rejects partial/unsafe structures and routes to clarification.

\item \textbf{Database (PostgreSQL):} Stores structured care record entries, including Resident identifiers, timestamps, category tags, and free-text statements, providing a reliable foundation for retrieval and auditing.

\item \textbf{Retrieval and question-answering component:} Retrieves relevant records using sparse retrieval (e.g.\ BM25), dense retrieval (vector similarity), and hybrid approaches, accommodating variation in how care staff phrase spoken queries. If retrieval confidence is weak (e.g., low similarity / conflicting evidence), the system either asks a disambiguation question or returns ``insufficient evidence'' rather than guessing.

\item \textbf{Scheduling component and calendar integration:} Converts extracted reminders into scheduled events and integrates with a calendar system. This separation of ``understanding the request'' from ``executing the task'' enables safety checks before actions are taken.

\item \textbf{Interaction and confirmation component:} Provides spoken responses, confirmations, and follow-up prompts, helping to reduce errors caused by ambiguous phrasing, partial speech, or uncertainty in timing details.
\end{itemize}

\subsection{Design choices supporting safety and reliability}

Because care homes are safety-critical environments, the system is designed to prioritise safety and reliability over full automation. Three design choices underpin this approach. First, confidence-aware behaviour identifies low-certainty outputs, which are flagged rather than acted upon automatically. Second, clarification before execution is employed when input is incomplete or ambiguous, reducing the risk of incorrect assumptions. Third, human-in-the-loop oversight is retained for actions that could plausibly affect resident safety, with confirmation requested before proceeding.

By structuring the system as an end-to-end pipeline with explicit safety checks at multiple points, the Care Home Smart Speaker is designed to support practical care workflows while maintaining transparency and control. This architectural design provides the foundation for the evaluation framework described in the following sections.

\section{Problem Definition and Evaluation Objectives}

Voice-enabled systems used in care homes operate within safety-critical environments, where errors may directly affect vulnerable individuals. Unlike consumer smart speakers, systems that support care documentation, information retrieval, and reminders must function reliably in noisy environments, with diverse accents, informal speech, and time pressure~\cite{koenecke2020racial, adedeji2024sound}. These challenges are compounded when systems are implemented as multi-stage pipelines, in which errors introduced at one stage may propagate downstream without being immediately visible~\cite{raghu2020survey}. A key limitation of existing evaluations of voice-based care technologies is their tendency to focus on isolated components or short-term usability outcomes~\cite{laranjo2018conversational}. For care workflows, this is insufficient because the safety outcome depends on the end-to-end behaviour. A “good” ASR model does not guarantee correct structured records, and correct records do not guarantee safe reminder scheduling or grounded retrieval responses, as evidenced by early-stage trials with care providers, which indicate that failures often arise not from a single component but from interactions among the transcription, parsing, retrieval, and task-execution stages~\cite{bates2021potential}. For example, during user testing with local care workers, transcription errors or ambiguous phrasing led to incorrect or missed retrieval outcomes, even though individual components functioned as intended. This motivates the need for an evaluation framework that treats voice-enabled care systems as end-to-end, safety-critical pipelines, rather than collections of independent modules.

\subsection{Conceptualising the system as a pipeline}

The Care Home Smart Speaker has been conceptualised as a sequence of interacting components that transform spoken input into structured care records and, where appropriate, into scheduled actions. Spoken input is transcribed, parsed into structured data, stored, retrieved in response to queries, and used to trigger reminders or notifications. Each stage introduces uncertainty, and system behaviour depends on how these uncertainties accumulate and are managed across the pipeline~\cite{raghu2020survey, sendak2020path}. Insights from early user testing trials with two care homes (NLG Care Home and Beverley Grange Nursing Home
) highlighted that real-world usage introduces variability not captured by laboratory-style testing. Differences in accents, unfamiliarity with the system, and environmental noise affected recognition and retrieval accuracy, even when the underlying models performed well under controlled conditions~\cite{koenecke2020racial}. As a result, evaluation in this paper considers not only average performance but also how the system behaves under uncertainty~\cite{amershi2019software}.

Furthermore, two constraints are particularly important in safety-critical care settings. First, the cumulative effect of errors across stages must remain within acceptable limits, as small inaccuracies may combine to produce significant downstream consequences~\cite{sendak2020path}. Similar concerns have been highlighted in research on safety assurance for AI-enabled systems, where failures in complex pipelines may propagate across components and undermine system safety~\cite{burton2020mind, calinescu2017engineering}. Second, the system must operate within bounded time constraints, as delayed responses may undermine the usefulness of reminders or real-time information access. To support systematic evaluation, we formalise the system as a pipeline consisting of $n$ interacting agents. The system can therefore be represented as a set of components $ S = \{A_1, A_2, \ldots, A_n\} $ where each agent $A_i$ represents a processing stage within the pipeline, such as speech recognition, natural language parsing, database insertion, retrieval, or reminder scheduling. Each agent receives an input $x_i$ and produces an output $y_i$ according to a potentially stochastic transformation:

\begin{equation}
y_i = A_i(x_i) + \epsilon_i
\label{eq:agent_error}
\end{equation}

where $\epsilon_i$ represents the error introduced by the $i$-th component, which may arise from transcription errors, parsing ambiguity, or retrieval inaccuracies~\cite{bishop2006pattern}. The overall output of the system can therefore be expressed as the composition of these agents:

\begin{equation}
Y = A_n \circ A_{n-1} \circ \cdots \circ A_1 (X) + \sum_{i=1}^{n} \epsilon_i
\label{eq:pipeline_composition}
\end{equation}

Equation~\ref{eq:pipeline_composition} captures the sequential nature of the pipeline, where the output of one stage becomes the input to the next stage. Similar formulations are commonly used to describe multi-stage machine learning systems in which errors introduced at early stages may propagate through downstream components~\cite{amershi2019software, raghu2020survey}. As a result, failures in upstream components such as speech recognition may influence later processes, including parsing, retrieval, or scheduling. 

Additionally, in the context of care delivery, another key concern is whether the cumulative error across the pipeline remains within an acceptable threshold. We denote the expected cumulative error as:

\begin{equation}
E = \mathbb{E}\left[\sum_{i=1}^{n} \epsilon_i \right]
\label{eq:cumulative_error}
\end{equation}

As expressed in Equation~\ref{eq:cumulative_error}, the expected error represents the aggregated uncertainty across all pipeline stages. The use of expectation operators to model aggregated stochastic error terms is standard in probabilistic machine learning and statistical modelling~\cite{bishop2006pattern}. To ensure safe operation in care environments, the system must satisfy the constraint as $ E \leq \delta $. It defines an acceptable upper bound on cumulative system error. In safety-critical AI applications such as healthcare, establishing tolerable error thresholds is essential for ensuring that automated systems operate within safe limits~\cite{bates2021potential}. In addition to accuracy constraints, care applications also impose latency constraints on system responses. Let $t_i$ denote the processing time of the $i$-th agent. The total system response time can therefore be defined as:

\begin{equation}
T = \sum_{i=1}^{n} t_i
\label{eq:latency_sum}
\end{equation}

Equation~\ref{eq:latency_sum} models the cumulative latency of the pipeline, reflecting the sequential processing of multiple system components. Similar latency aggregation models are commonly used in distributed and multi-component computing systems~\cite{van2002distributed}. For real-time interactions, such as spoken queries or scheduled reminders, the total response time must satisfy the constraint $ T \leq \tau $. It ensures that system responses remain within acceptable timing bounds required for practical care workflows. Together, the above equations highlight that evaluating individual components in isolation is insufficient for safety-critical care systems. Even if each agent performs well independently, the combined effect of small errors or delays may lead to unacceptable system behaviour~\cite{amershi2019software}. The evaluation problem addressed in this study is therefore defined as the construction of an evaluation framework $F$ that, for a given set of inputs $X$ representing realistic care interactions, measures:

\begin{itemize}
\item Agent-level performance metrics $M(A_i)$, such as transcription accuracy or parsing correctness;
\item System-level reliability $R(S)$, capturing end-to-end behaviour under noise, ambiguity, and challenging inputs;
\item Trust-enabling properties $\Theta(S)$, including confidence scoring, clarification prompts, and human-in-the-loop verification~\cite{bates2021potential}.
\end{itemize}

Formally, the evaluation framework is expressed as $ F : S \times X \rightarrow (M, R, \Theta) $ and it defines the mapping from the system and its input space to a set of evaluation outcomes. Such framework-based evaluation approaches are widely used in assessing the reliability and safety of complex AI systems deployed in healthcare and other high-risk environments~\cite{sendak2020path}. This formulation directly informs the evaluation framework and metrics described in the following section.

\section{Evaluation Framework and Metrics}

Evaluating a voice-enabled system for care homes requires more than reporting isolated performance numbers. In safety-critical care settings, the key issue is whether the system can be shown using transparent evidence to behave reliably across the full pipeline (speech → parsing → storage → retrieval → reminders), and to manage uncertainty safely when it arises. To address this, an assurance-driven evaluation framework is used that links risks to measurable claims and trial evidence~\cite{bates2021potential, sendak2020path}. The metrics cover the full range of system functions: ensuring that voice inputs are accurately transcribed, that structured data is correctly inserted into the database, and that reminders are validated and appropriately scheduled. We also measured how well the Large Language Model (LLM) parsing and Retrieval-Augmented Generation (RAG) methods handled different types of queries, including paraphrased and ambiguous ones. Particular focus was placed on matching the correct Resident ID and care category, validating reminders, and maintaining consistency in information retrieval, all essential for avoiding mistakes and safeguarding resident wellbeing~\cite{lewis2020retrieval, belisle2024stakeholder}.

\subsection{Assurance-driven evaluation approach}
\label{sec:parsingrisk}
\begin{figure}[H]
  \centering
  \includegraphics[width=1\linewidth,keepaspectratio]{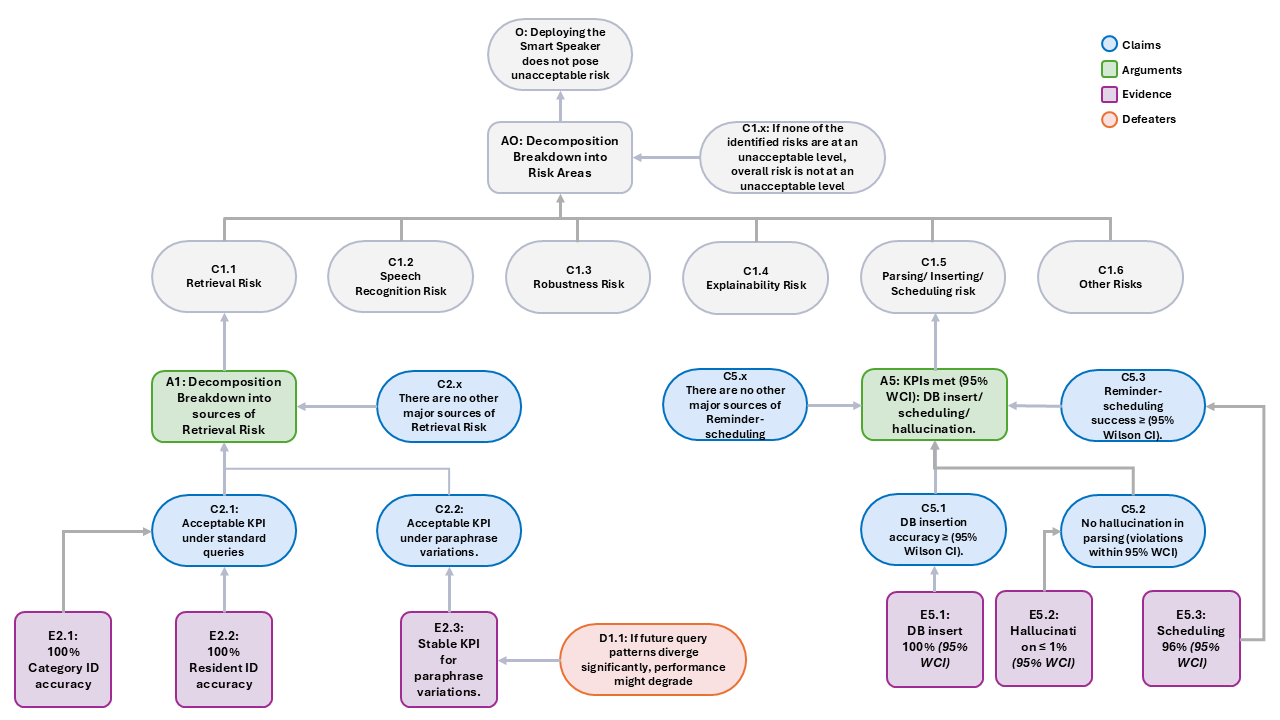}
  \caption{
  Assurance case for the Care Home Smart Speaker.
  The metrics-based argument \textbf{A2} justifies the parsing, inserting, and scheduling risk \textbf{C1.5} using three KPIs evaluated with 95\% Wilson confidence intervals:
  \textbf{C2.3} database insertion accuracy,
  \textbf{C2.4} reminder scheduling success,
  and \textbf{C2.5} absence of hallucinations during parsing.
  Retrieval risk (\textbf{C1.1}) is decomposed by \textbf{A1} into standard and paraphrased queries, supported by evidence \textbf{E2.1--E2.3} and defeater \textbf{D2.2}.
  }
  \label{fig:assurance}
\end{figure}

Evaluation is structured around a safety-style assurance argument: a top-level claim is broken down into specific risk areas (e.g., speech recognition risk, retrieval risk, and parsing/scheduling risk), and each risk area is supported by empirical evidence. Assurance cases are commonly used in safety-critical domains to demonstrate that system risks have been systematically identified, mitigated, and supported by evidence~\cite{kelly2004goal, hawkins2021guidance}. Figure~\ref{fig:assurance} presents the overall assurance case used to structure this evaluation. The figure illustrates how the top-level claim that deployment of the Care Home Smart Speaker does not pose unacceptable risk is decomposed into distinct risk areas, including speech recognition, retrieval, robustness, explainability, and parsing, insertion, and scheduling. As shown in Figure~\ref{fig:assurance}, this claim is decomposed into specific risk areas, including retrieval risk (C1.1), speech recognition risk (C1.2), robustness risk (C1.3), explainability risk (C1.4), and parsing, insertion, and scheduling risk (C1.5). Each of these areas is evaluated separately, while also considering how risks interact across the end-to-end pipeline. Each risk area is supported by arguments and evidence derived from quantitative evaluation and observations from care-home trials. This representation makes explicit how evaluation metrics are used not merely to report performance, but to justify safety-relevant claims~\cite{bates2021potential}.

\section{Evaluation dimensions}

Guided by the assurance case, the system is evaluated across four core dimensions:

\begin{enumerate}

\item \textbf{Speech recognition reliability:}
Speech recognition reliability addresses the risk that spoken input may be incorrectly transcribed due to background noise, accent diversity, or informal phrasing. Evaluation focuses not only on transcription accuracy but also on whether real-time safeguards prevent low-quality audio from producing misleading outputs. Speech recognition performance is widely recognised as a critical factor in voice-enabled healthcare systems where transcription errors may affect downstream clinical decision-making~\cite{koenecke2020racial, belisle2024stakeholder}. In the assurance case (Figure~\ref{fig:assurance}), evidence supporting this dimension contributes to mitigating speech recognition risk (C1.2).

\item \textbf{Structured data integrity:}
Structured data integrity relates to the correct transformation of transcribed speech into valid care records. This includes accurate identification of residents, correct assignment of care categories, and successful insertion of records into the database. Errors at this stage may not be immediately visible but can influence downstream retrieval and task execution. Ensuring data integrity is a key requirement for AI systems deployed in healthcare environments where incorrect records may have safety implications~\cite{sendak2020path}. Evidence in this dimension supports arguments associated with parsing and insertion risk (C1.5).

\item \textbf{Retrieval correctness:}
This dimension ensures that relevant care information is surfaced when needed. The evaluation considers both standard queries and more challenging scenarios in which critical information is embedded in large volumes of routine documentation. Retrieval-augmented systems must be evaluated not only for accuracy but also for their ability to handle paraphrased or ambiguous queries~\cite{lewis2020retrieval}. As shown in Figure~\ref{fig:assurance}, retrieval performance evidence supports mitigation of retrieval risk (C1.1), including under paraphrased and non-standard query formulations.

\item \textbf{Task and reminder handling:}
Task and reminder handling focuses on the safe extraction, scheduling, and execution of time-based instructions. This includes evaluating whether reminders are scheduled correctly, whether ambiguity triggers clarification, and whether incorrect or low-confidence actions are avoided. Such safeguards are particularly important for AI systems deployed in safety-critical healthcare contexts~\cite{bates2021potential}. This dimension directly addresses parsing, inserting, and scheduling risk (C1.5), which is explored in more detail in Section~\ref{sec:parsingrisk}.
\end{enumerate}

\subsection{Data integrity and retrieval metrics}

A core part of evaluating the Care Home Smart Speaker is ensuring it can reliably store and retrieve information in real-world care settings. This involves confirming that voice inputs are transcribed accurately, that care records are correctly parsed and stored, and that reminders are validated and described consistently. Database integrity and reliable information retrieval are widely recognised requirements for trustworthy healthcare information systems~\cite{sendak2020path}. First, database accuracy was assessed to determine whether resident care records, transcribed from staff speech, were correctly parsed by the LLM and inserted into the PostgreSQL database without loss or corruption of critical details. For example, when a care worker recorded a note about medication or equipment, the system was expected to capture and log the details precisely in the correct fields.

\subsubsection{Resident ID and category accuracy}

Equally important was the system’s ability to retrieve the correct information. Metrics based on correct classification proportions are commonly used to evaluate structured information extraction tasks in NLP systems~\cite{jurafsky2000speech}. To measure this, Matching Resident ID and Category Accuracy was evaluated, which ensures that the correct resident and the intended care category (e.g., mobility, medication, or personal hygiene) are retrieved. This metric is particularly important in care contexts because even a small mismatch may associate information with the wrong person, which is unacceptable in practice. This metric is formally defined in Equation~\ref{eq:id_cat_accuracy}.

\begin{equation}
\mathrm{Accuracy}_{\text{ID+Category}} =
\frac{\text{Number of queries with both correct resident ID and category}}
{\text{Total number of queries}}
\label{eq:id_cat_accuracy}
\end{equation}

\subsubsection{Semantic reliability (meaning preservation)}

Accuracy alone does not guarantee that the system preserves meaning, especially when staff phrase reminders in different ways. For this reason, embedding-based measures are also included for evaluation. One of the measures has been Cosine similarity to assess, as Cosine similarity is widely used in information retrieval and semantic search tasks~\cite{schutze2008introduction}. The Cosine similarity measures how closely the query aligns semantically with stored record embeddings (higher values indicate stronger similarity), as shown in Equation~\ref{eq:cosine}. 
\begin{equation}
\mathrm{CosineSimilarity}(x, y) =
\frac{x \cdot y}{\|x\| \, \|y\|}
\label{eq:cosine}
\end{equation}

Complementing this, Word Mover’s Distance (WMD) is another measure that is used to capture semantic distance between retrieved text and ground-truth descriptions. WMD measures the minimal cost required to transform one document into another in embedding space~\cite{kusner2015word}. The metric is defined in Equation~\ref{eq:wmd}.

\begin{equation}
\mathrm{WMD}(D_1, D_2) =
\min_{T \geq 0}
\sum_{i,j} T_{ij} \, \| w_i - w_j \|
\label{eq:wmd}
\end{equation}

subject to flow constraints that preserve the normalised word distributions of documents $D_1$ and $D_2$. Lower WMD values indicate closer semantic alignment.

\subsubsection{Statistical confidence (Wilson confidence intervals)}

To quantify statistical confidence in observed accuracy rates, 95\% Wilson confidence interval is applied. The Wilson interval provides improved coverage accuracy compared to the normal approximation when dealing with binomial proportions~\cite{wilson1927probable}. The interval is defined in Equation~\ref{eq:wci}.

\begin{equation}
CI =
\frac{\hat{p} + \frac{z^2}{2n} \pm z
\sqrt{\frac{\hat{p}(1 - \hat{p})}{n} + \frac{z^2}{4n^2}}}
{1 + \frac{z^2}{n}}
\label{eq:wci}
\end{equation}

where $\hat{p}$ is the observed proportion, $n$ is the sample size, and $z = 1.96$.

Together, these measures ensure that the Care Home Smart Speaker not only stores care records accurately but also retrieves them reliably while preserving semantic meaning and supporting safe, trustworthy care.

\subsection{Formal Evaluation Metrics}

Accuracy measures the overall proportion of correct predictions across all evaluated instances. It is formally defined in Equation~\ref{eq:accuracy}. Precision and recall provide complementary insights into system behaviour and are widely used evaluation metrics for classification tasks~\cite{powers2020evaluation}. They are defined in Equations~\ref{eq:precision} and~\ref{eq:recall}. Collectively, these Equations quantify correctness, robustness, and responsiveness. Reporting both point estimates and confidence intervals enables statistically grounded assessment of safety and reliability in real-world care environments.
\begin{equation}
\mathrm{Accuracy} = \frac{TP + TN}{TP + TN + FP + FN}
\label{eq:accuracy}
\end{equation}

\begin{equation}
\mathrm{Precision} = \frac{TP}{TP + FP}
\label{eq:precision}
\end{equation}

\begin{equation}
\mathrm{Recall} = \frac{TP}{TP + FN}
\label{eq:recall}
\end{equation}





\section{Experimental Setup and Result Analysis}

This section describes the experimental setup used to evaluate the Care Home Smart Speaker. The setup was designed to balance methodological rigour with the practical and ethical constraints of evaluating a safety-critical system intended for use in care environments. Evaluation, therefore, combined supervised care-home trials with controlled testing, allowing system behaviour to be examined under realistic conditions without introducing risk to residents. The experimental design directly supports the evaluation objectives defined in Sections~4 and~5 and provides the empirical evidence used in the assurance case. Table~\ref{tab:experimental_setup} summarises the experimental configuration used to evaluate the Care Home Smart Speaker across its main functional components. The evaluation combined supervised care-home trials with controlled testing to assess system behaviour under realistic operational conditions. A total of 330 spoken interactions were collected during pilot trials conducted in two residential care homes, covering routine care documentation, information queries, and reminder-related tasks. From these interactions, 184 transcripts contained reminder-related content and were used to evaluate reminder extraction and scheduling performance. Each stage of the system pipeline was assessed independently, including speech recognition, structured data parsing and database insertion, information retrieval under varied query formulations, and reminder scheduling. For each component, the table specifies the configuration used during testing, the data source and number of evaluation instances, and the system outputs logged for subsequent quantitative analysis. Results are presented for structured data integrity, retrieval performance, and reminder scheduling. All results are derived from logged system outputs generated during supervised care-home trials and controlled testing.

\begin{table*}[hpt!]
\centering
\caption{Summary of the experimental setup used to evaluate the Care Home Smart Speaker system.}
\label{tab:experimental_setup}

\begin{tabular}{p{2.2cm} p{4.2cm} p{2.5cm} p{2cm} p{3cm}}
\hline
\hline
\textbf{Evaluation Component} & \textbf{Configuration / Scenario} & \textbf{Data Source} & \textbf{No. of Instances} & \textbf{Logged Outputs} \\
\hline
Care-home trial context &
Supervised pilot trials conducted in two UK residential care homes (NLG and Beverley Grange, Hull). Interactions occurred in typical staff areas with background noise and interruptions. The system operated strictly as a decision-support tool. &
Care-home staff interactions &
2 sites &
Trial logs, qualitative observations \\

Participants and interaction scenarios &
Care staff interacted with the system using natural speech. Scenarios included care documentation, note recording, care information queries, and reminder scheduling. Both scripted and unscripted interactions were included. &
Spoken interactions &
330 transcripts &
Audio recordings, transcripts, parsed outputs \\

Speech recognition and preprocessing &
Whisper-based ASR with signal-quality safeguards and rejection of low-confidence audio segments to prevent unreliable transcriptions. &
Spoken input &
330 audio inputs &
Accepted/rejected audio segments, ASR transcripts \\

Parsing and database insertion &
Transcripts parsed into structured records containing resident ID, care category, timestamp, and free-text notes. Schema validation enforced before insertion into a PostgreSQL database. &
ASR transcripts &
330 parsed records &
Structured database entries, validation logs \\

Reminder extraction &
Reminder descriptions extracted from spoken notes when reminder intent was detected. &
Reminder-containing transcripts &
184 cases &
Extracted reminder text, classification outputs \\

Retrieval evaluation &
Spoken queries issued by care staff and evaluators using sparse, dense, and hybrid retrieval strategies. Needle-in-a-haystack tests embedded critical records within larger document sets. &
Trial + synthetic database &
5 representative queries &
Retrieved records, ranking positions, similarity scores \\

Reminder scheduling and task handling &
Extracted reminders processed by the scheduling module and added to a calendar system when confidence thresholds were satisfied. Background scheduler triggered reminders during test runs. &
Reminder cases &
184 scheduling attempts &
Scheduled events, triggered reminders, clarification prompts \\

Qualitative interaction observations &
Supervised observations of interaction behaviour including clarification requests, confusion events, and dialogue breakdowns. &
Trial sessions &
All sessions &
Observation notes \\

Ethical and governance safeguards &
Evaluation conducted under supervised user testing. No real resident data used. All stored records were synthetic or anonymised. &
Entire evaluation &
N/A &
Ethics documentation \\

\hline
\hline
\end{tabular}
\end{table*}

\subsection{Accuracy of Inserting Data into the Database}

A total of 330 transcripts were evaluated, spanning 11 care categories, with three residents represented in each category. Among these transcripts, 184 contained reminder-related content, enabling evaluation of both structured database insertion and reminder extraction. The assessment focused on three tasks: (i) correct matching of care categories, (ii) correct identification of resident IDs, and (iii) accurate recognition of reminder descriptions. As shown in Table~\ref{tab:db_accuracy_models}, model performance varied across the three tasks. For structured database insertion, GPT~5.2 achieved perfect accuracy (100\%) for both care category matching and resident ID identification, with tight 95\% Wilson confidence intervals (98.86\%--100\%), indicating highly reliable structured extraction. LLaMA--3 also demonstrated strong performance, achieving 97.87\% accuracy for both tasks. In contrast, Qwen exhibited substantially lower accuracy, achieving 67.88\% category matching accuracy (95\% CI: 62.66\%--72.69\%) and 66.36\% resident ID matching accuracy (95\% CI: 61.10\%--71.25\%).

Importantly, note insertion was fully dependent on the correct identification of both the care category and resident ID. Under these conditions, note information was consistently inserted without error, resulting in 100\% note insertion accuracy whenever the corresponding structured fields were correctly identified. Reminder recognition proved to be the most challenging task across all models. GPT~5.2 achieved a reminder recognition accuracy of 89.09\% (95\% CI: 83.81\%--92.80\%), indicating strong but non-perfect performance in extracting reminder descriptions. LLaMA--3 achieved a reminder recognition accuracy of 71.19\% (95\% CI: 64.10\%--77.37\%), while Qwen performed less robustly with an accuracy of 58.15\% (95\% CI: 50.93\%--65.04\%). These results suggest that reminder extraction is more sensitive to linguistic variability, implicit phrasing, and contextual ambiguity than structured fields such as care category or resident ID, particularly for smaller models.

\begin{table}[h]
\centering
\caption{Database insertion and reminder recognition accuracy across models with 95\% Wilson confidence intervals.}
\label{tab:db_accuracy_models}
\begin{tabular}{lcccccc}
\hline
\textbf{Metric} & \textbf{Model} & \textbf{Accuracy (\%)} & \textbf{CI Lower (\%)} & \textbf{CI Upper (\%)} \\
\hline
Category Match       & GPT 5.2 &   \textbf{100} & 98.86 & 100\\
                     & LLaMA--3 & 97.87   & 95.77 & 99.08 \\
                     & Qwen     &  67.88 & 62.66 & 72.69\\
Resident ID Match    & GPT 5.2 &  \textbf{100}  & 98.86 & 100  \\
                     & LLaMA--3 & 97.87   & 95.68 & 98.96  \\
                     & Qwen     &  66.36  & 61.10 & 71.25 \\
Reminder Recognition & GPT 5.2 & \textbf{89.09} & 83.81 & 92.80  \\
                     & LLaMA--3 & 71.19 & 64.10 & 77.37 \\
                     & Qwen     & 58.15 & 50.93 & 65.04  \\
\hline
\end{tabular}

\end{table}

\noindent\textbf{Reminder Recognition Error Analysis.}
To further analyze reminder recognition behavior, Table~\ref{tab:reminder_confusion_models} reports the confusion matrices for each model. The task is formulated as a binary classification problem, where the objective is to predict whether a transcript contains a reminder (positive class) against ground-truth labels. For GPT~5.2, the results (TP~=~184, FP~=~36, FN~=~0, TN~=~110) indicate that all reminder-containing transcripts were successfully detected, yielding zero missed reminders and corresponding to 100\% recall. This strong recall performance is accompanied by a limited number of false positives, suggesting a tendency to conservatively flag potential reminders in linguistically ambiguous cases. In contrast, both LLaMA--3 and Qwen exhibited no false positives (FP~=~0) but substantially higher false negative rates, missing 53 and 77 reminders, respectively. This pattern reflects a more conservative prediction strategy that prioritizes precision over recall, resulting in lower overall reminder recognition accuracy.

\begin{table}[h]
\centering
\caption{Confusion matrix (TP/FP/FN/TN) for reminder recognition across models.}
\label{tab:reminder_confusion_models}
\begin{tabular}{lcccc}
\hline
\textbf{Model} & \textbf{TP} & \textbf{FP} & \textbf{FN} & \textbf{TN} \\
\hline
GPT 5.2  & 184 & 36 & 0 & 110 \\
LLaMA--3 & 131  & 0 & 53 & 146  \\
Qwen     & 107  & 0 & 77 & 146  \\
\hline
\end{tabular}

\end{table}

Notably, a subset of the evaluated transcripts intentionally contained misleading or ambiguous expressions to assess model robustness under realistic conditions. While such cases had minimal impact on structured field insertion (care category, resident ID, and notes), they disproportionately affected reminder recognition performance, underscoring the importance of challenging examples for stress-testing reminder extraction systems.

\subsection{Semantic Fidelity of Extracted Reminders and Notes}

Beyond accuracy-based metrics, the semantic closeness between system-generated text and the corresponding ground-truth descriptions is also evaluated, using Cosine Distance and Word Mover’s Distance (WMD). These distance-based metrics assess whether the underlying meaning of reminders and notes is preserved, even when surface-level phrasing differs from the reference text. As shown in Table~\ref{tab:text_distance_summary_models}, GPT~5.2 produced the most semantically aligned reminder descriptions, achieving the lowest mean cosine distance (0.0743) and WMD (0.3242) across 184 reminder instances. LLaMA--3 and Qwen exhibited progressively higher distances, with mean WMD values of 0.4236 and 0.5476, respectively, indicating reduced semantic fidelity in reminder extraction. A similar trend is observed for care notes. GPT~5.2 consistently maintained lower cosine distance and WMD values than both LLaMA--3 and Qwen, suggesting more reliable preservation of meaning during note extraction and database insertion. Overall, these findings reinforce that while structured data insertion is largely robust, semantic understanding remains a key challenge for reminder recognition, particularly under ambiguous or misleading linguistic conditions.

\begin{table}[h]
\centering
\caption{Distance summary between ground-truth and system-generated text. Lower values indicate higher semantic similarity. P25 and P75 denote the 25th and 75th percentiles.}
\label{tab:text_distance_summary_models}
\begin{tabular}{llccccc}
\hline
\textbf{Field} & \textbf{Metric} & \textbf{Model} & \textbf{n} & \textbf{Mean} & \textbf{P25} & \textbf{P75} \\
\hline
Reminder & Cosine Distance & GPT 5.2     & 184 & 0.0743 & 0.0274 & 0.1022  \\
Reminder & Cosine Distance & LLaMA--3 & 127 & 0.1031  & 0.0406    & 0.1410  \\
Reminder & Cosine Distance & Qwen     & 106  & 0.1630    & 0.0989     & 0.2013     \\

Reminder & WMD             & GPT 5.2     & 184 & 0.3242 & 0.2059 & 0.4460  \\
Reminder & WMD             & LLaMA--3 & 127  & 0.4236  & 0.2462     & 0.5690     \\
Reminder & WMD             & Qwen     & 106  & 0.5476     & 0.4266     & 0.6557     \\
\hline
Notes    & Cosine Distance & GPT 5.2     & 330 & 0.0829 & 0.0618 & 0.0978 \\
Notes    & Cosine Distance & LLaMA--3 & 323  & 0.2296     & 0.1857   & 0.2663     \\
Notes    & Cosine Distance & Qwen     & 222  & 0.2006     & 0.1557     & 0.2427     \\

Notes    & WMD             & GPT 5.2     & 330 & 0.5476 & 0.4632 & 0.6302 \\
Notes    & WMD             & LLaMA--3 & 323  & 0.9002     & 0.8540     & 0.9601     \\
Notes    & WMD             & Qwen     & 222  & 0.8432     & 0.7731     & 0.9116     \\
\hline
\end{tabular}

\end{table}

\begin{figure}[h]
    \centering
    \includegraphics[width=1\linewidth]{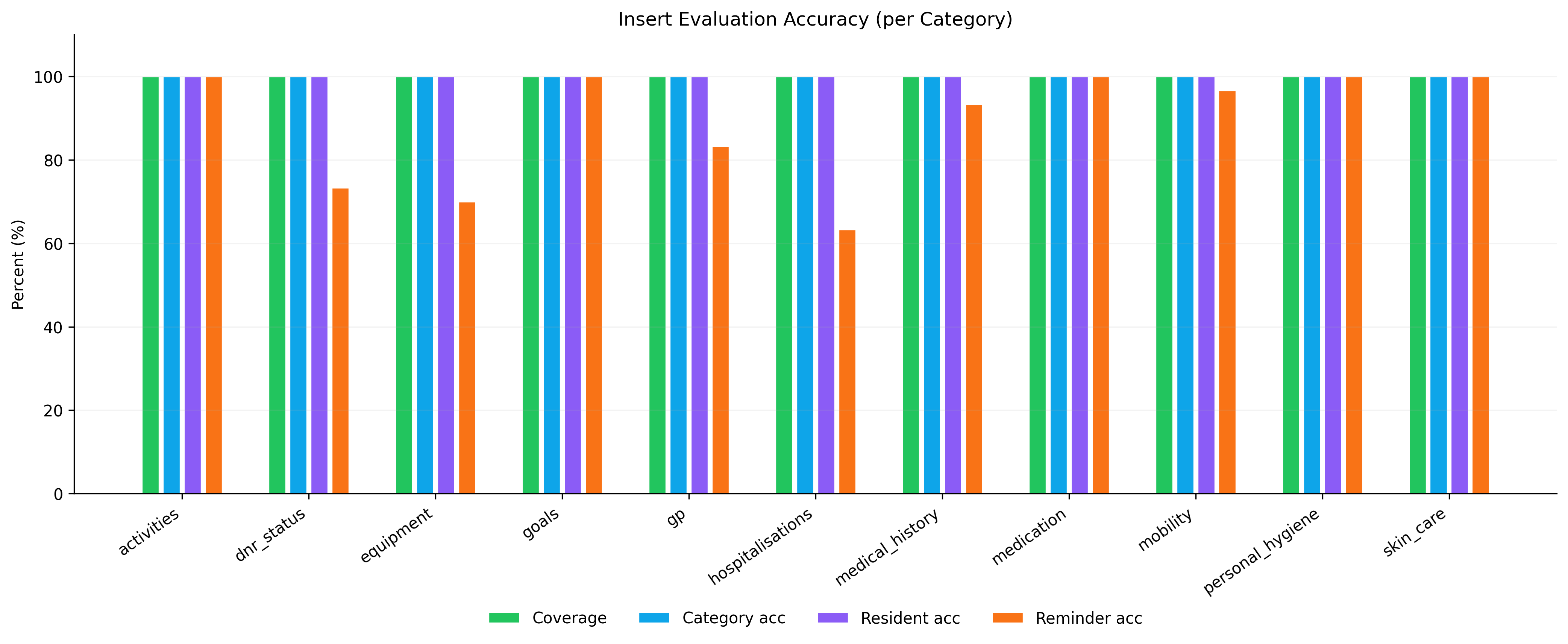}
    \caption{Per-category accuracy for GPT~5.2, showing category matching, resident ID matching, and reminder recognition across care categories.}

    \label{fig:gpt}
\end{figure}

\begin{figure}[h]
    \centering
    \includegraphics[width=1\linewidth]{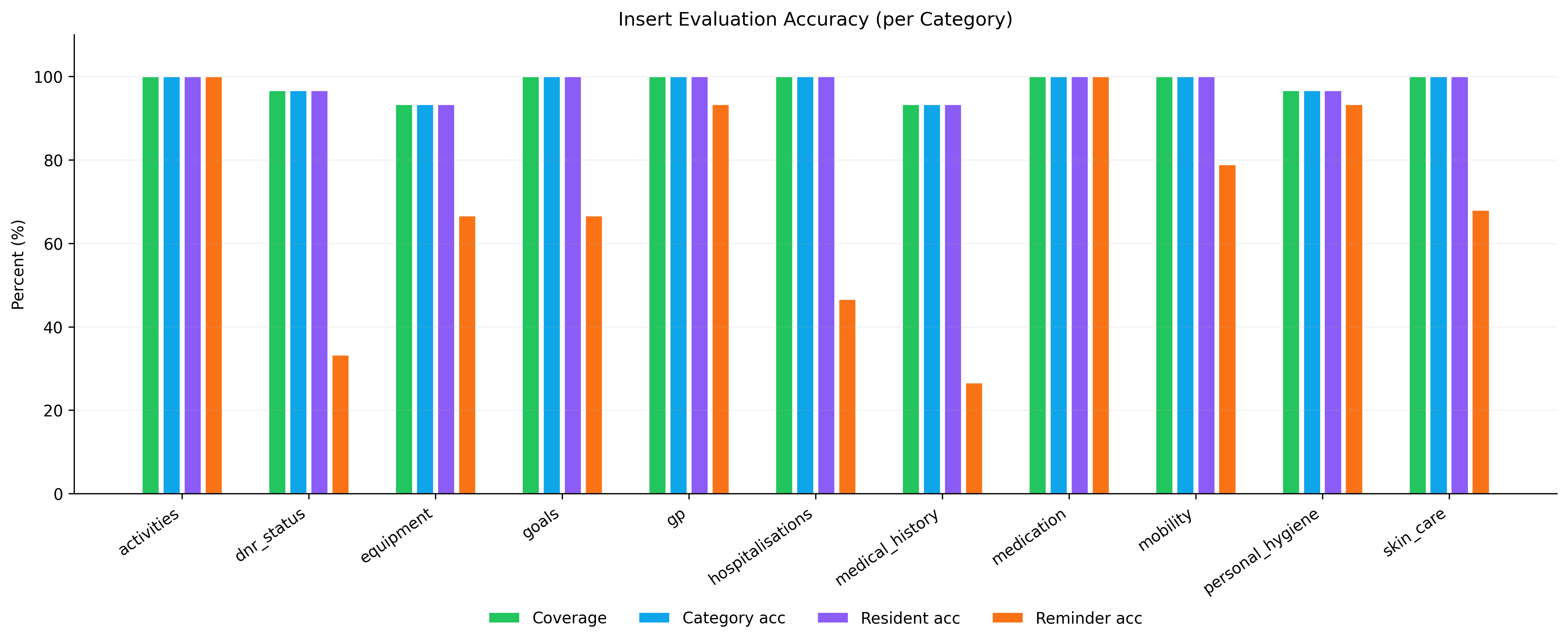}
    \caption{Per-category accuracy for LLaMA--3, showing category matching, resident ID matching, and reminder recognition across care categories.}

    \label{fig:llama}
\end{figure}

\begin{figure}[h]
    \centering
    \includegraphics[width=1\linewidth]{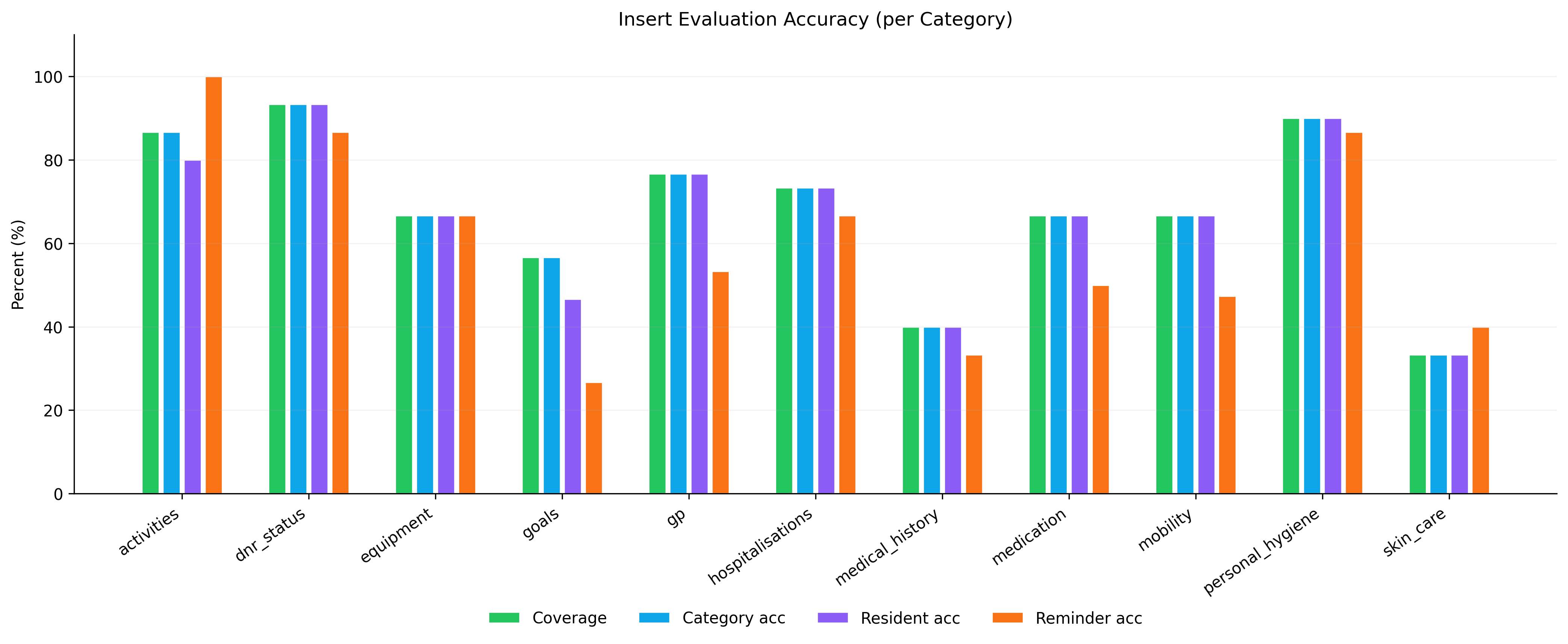}
    \caption{Per-category accuracy for Qwen, showing category matching, resident ID matching, and reminder recognition across care categories.}

    \label{fig:qwen}
\end{figure}

In addition to overall accuracy, Figures~\ref{fig:gpt}, \ref{fig:llama}, and \ref{fig:qwen} illustrate per-category performance for GPT~5.2, LLaMA--3, and Qwen, respectively. For each model, the figures report Category Accuracy, Resident ID Accuracy, and Reminder Accuracy across multiple care categories, including activities, equipment, goals, mobility, and personal hygiene. Across all three models, category and resident ID matching remain consistently high with limited variation across care categories. In contrast, reminder recognition exhibits greater variability between categories, reflecting differences in linguistic complexity and contextual ambiguity inherent to reminder expressions across care domains.

\clearpage

\subsection{Calendar Integration Performance}

Beyond correctly parsing and storing reminder information, the Care Home Smart Speaker was also evaluated on its ability to integrate with Google Calendar for reliable task scheduling. This step is critical in care environments, where timely reminders support both medication adherence and daily living activities. Table~\ref{tab:calendar_models} summarises the results of the calendar integration evaluation. For structured fields, category and resident ID matching again achieved perfect accuracy (100\%) for GPT~5.2, with tight 95\% Wilson confidence intervals, demonstrating the robustness of the LLM-based parsing pipeline when transferring reminders from transcripts into actionable calendar events. LLaMA--3 also achieved near-ceiling performance for these fields, while Qwen showed substantially lower accuracy, consistent with earlier database insertion results. For reminder count matching, GPT~5.2 achieved an accuracy of 84.65\% (95\% CI: 78.00\%--89.56\%), indicating that the majority of reminders were scheduled with the correct count, but with occasional mismatches. LLaMA--3 and Qwen performed less robustly on this task, both achieving reminder count matching accuracies of 55.68\%, reflecting greater difficulty in consistently translating parsed reminder information into the correct number of calendar entries.

\begin{table}[h]
\centering
\caption{Google Calendar integration accuracy with 95\% Wilson confidence intervals across models.}
\label{tab:calendar_models}
\begin{tabular}{lcccccc}
\hline
\textbf{Metric} & \textbf{Model} & \textbf{Accuracy (\%)} & \textbf{CI Lower (\%)} & \textbf{CI Upper (\%)} \\
\hline
Category Match       & GPT 5.2  & 100    & 97.95 & 100.00 \\
                     & LLaMA--3 & 98.18  & 96.09 & 99.16  \\
                     & Qwen     & 67.88  & 60.79 & 74.14  \\

Resident ID Match    & GPT 5.2  & 100    & 97.95 & 100.00 \\
                     & LLaMA--3 & 98.18  & 96.09 & 99.16  \\
                     & Qwen     & 66.36  & 59.25 & 72.79  \\

Reminder Count Match & GPT 5.2  & 84.65  & 78.00 & 89.56 \\
                     & LLaMA--3 & 55.68  & 45.84 & 65.11 \\
                     & Qwen     & 55.68  & 45.84 & 65.11 \\

\hline
\end{tabular}

\end{table}

Overall, these results confirm that the system can reliably schedule reminders and associate them with the correct resident and care category. The observed gap in reminder count matching highlights the need for further refinement in handling edge cases, such as closely spaced reminders, duplicate expressions, or ambiguous phrasing. Nevertheless, the calendar integration demonstrates a strong foundation for real-world deployment in care settings, enabling automated scheduling while maintaining safety and reliability.

\subsection{Resident Query Retrieval Accuracy}

To evaluate the system’s ability to retrieve relevant information in response to resident queries, retrieval performance using three retrieval strategies: Hybrid RAG, Sparse BM25, and Dense FAISS is assessed. This evaluation is conducted on a small set of five representative resident queries, and therefore, the results should be interpreted as indicative rather than statistically conclusive. Table~\ref{tab:rag_models} reports the mean semantic distance between retrieved responses and ground-truth answers using Cosine Distance and Word Mover’s Distance (WMD), with lower values indicating closer semantic alignment. Across retrieval methods, GPT~5.2 generally achieved lower retrieval distances than LLaMA--3 and Qwen, suggesting more semantically relevant retrieval under the tested conditions. Among the retrieval strategies, Sparse BM25 achieved the lowest cosine distances for GPT~5.2, while Hybrid RAG and Dense FAISS showed comparable performance across models. In terms of WMD, Hybrid RAG produced slightly lower distances for GPT~5.2 compared to the other methods, whereas LLaMA--3 and Qwen exhibited higher variability across retrieval strategies.

Overall, these results suggest that retrieval performance is sensitive to both the choice of retrieval method and the underlying language model. However, given the limited number of evaluation queries, these findings primarily serve as an initial comparison of retrieval behaviours rather than a definitive ranking. A larger and more diverse query set is required to draw stronger conclusions about retrieval robustness and generalisation in real-world care settings. 
Evidently, despite Sparse BM25 achieving lower cosine distances in some cases, qualitative inspection of the retrieved results indicated that Hybrid RAG generally produced more relevant and contextually appropriate responses. This observation suggests that cosine distance alone may not be a fully reliable indicator of retrieval quality for natural language responses, as it can favour lexical similarity over semantic usefulness. In contrast, Hybrid RAG appeared more effective at capturing contextual and semantic relationships relevant to resident queries, highlighting the importance of complementary qualitative analysis when evaluating retrieval performance.

\begin{table}[ht]
\centering
\caption{Mean retrieval distances across RAG methods per model. Lower distances indicate closer alignment with ground truth.}
\label{tab:rag_models}
\begin{tabular}{llcc}
\hline
\textbf{Method} & \textbf{Metric} & \textbf{Model} & \textbf{Mean} \\
\hline
Hybrid RAG  & Cosine Distance & GPT 5.2   & 0.371541 \\
            &                 & LLaMA--3  & 0.428666 \\
            &                 & Qwen      & 0.418436 \\
Hybrid RAG  & WMD Distance    & GPT 5.2   & 0.743192 \\
            &                 & LLaMA--3  & 0.883987 \\
            &                 & Qwen      & 0.950280 \\
Sparse BM25 & Cosine Distance & GPT 5.2   & 0.308598 \\
            &                 & LLaMA--3  & 0.507080 \\
            &                 & Qwen      & 0.346649 \\
Sparse BM25 & WMD Distance    & GPT 5.2   & 0.794588 \\
            &                 & LLaMA--3  & 0.968074 \\
            &                 & Qwen      & 0.853969 \\
Dense FAISS & Cosine Distance & GPT 5.2   & 0.401626 \\
            &                 & LLaMA--3  & 0.394066 \\
            &                 & Qwen      & 0.408740 \\
Dense FAISS & WMD Distance    & GPT 5.2   & 0.751201 \\
            &                 & LLaMA--3  & 0.847594 \\
            &                 & Qwen      & 0.904435 \\
\hline
\end{tabular}

\end{table}

\section{Discussion and Limitations }

This study evaluated a pipeline-based (multi-component), voice-enabled smart speaker system designed to support care documentation, information retrieval, and reminder scheduling in care home settings. The evaluation combined supervised trials conducted in two UK care homes with controlled testing, using an assurance-driven framework to examine system behaviour across the full processing pipeline. This section discusses the findings in relation to the evaluation objectives and outlines the study's limitations. Throughout, we interpret performance in terms of safety-relevant outcomes, distinguishing unsafe failures (e.g., wrong resident attribution, missed reminders, incorrect schedules) from safe deferrals (e.g., clarification prompts or human confirmation).

\subsection{Discussion}
The results demonstrate that the system converted spoken interactions into structured care records with high accuracy in resident identifier matching, care category assignment, and database insertion. These outcomes indicate that spoken input can be transformed into structured representations suitable for storage and later retrieval under supervised trial conditions. This strong performance is most plausible for constrained structured fields, where parsing is bounded by a fixed schema and validated against required fields, reducing the space for uncontrolled outputs.
Reminder-related results show that a majority of spoken reminder requests were successfully scheduled and executed, with a smaller proportion not resulting in scheduling due to insufficient extracted information. This highlights the practical challenge of handling informal and sometimes incomplete spoken instructions in real care environments, where staff may not always specify precise timing details. A key safety-relevant observation in this work is the trade-off between missed reminders (false negatives) and spurious reminders (false positives): configurations that avoid missed reminders can still over-trigger on reminder-like phrasing. In care contexts, missed reminders may carry higher potential harm, whereas false positives can increase workload and contribute to “alert fatigue” if not gated. These findings support a conservative operational policy where reminders are treated as suggested intents and require explicit confirmation before scheduling when confidence is low or temporal expressions are underspecified. Retrieval performance results indicate that the system returned relevant records for a majority of spoken queries, with overall retrieval performance varying by retrieval strategy and model. A comparative evaluation across sparse, dense, and hybrid retrieval strategies suggests that hybrid retrieval can be advantageous under paraphrased or non-standard phrasing, consistent with the expectation that care staff may phrase queries in diverse ways that benefit from combining lexical and semantic retrieval. Across evaluation dimensions, results were reported with Wilson confidence intervals, providing an explicit representation of uncertainty associated with proportion-based metrics. The use of confidence intervals supports cautious interpretation of performance, particularly given the scale and scope of the trials. Where distance-based semantic metrics are used, they should be interpreted as proxy measures of alignment rather than definitive “accuracy,” and distributional summaries (e.g., percentiles) are valuable for understanding worst-case behaviour in safety-critical settings.

\subsection{Implications for voice-enabled systems in care}

Taken together, the findings suggest that voice-enabled systems can support structured data capture, retrieval, and reminder-related tasks under supervised care-home trial conditions. The results also highlight that system performance varies across different stages of the pipeline, and that downstream outcomes depend on the accuracy and completeness of earlier processing stages. The evaluation framework used in this study emphasises end-to-end behaviour rather than isolated component performance. This perspective is important in care settings, where minor errors introduced early in the pipeline may have disproportionate downstream effects. The results underscore the need for evaluation approaches that explicitly account for error propagation, uncertainty, and interactions among system components.
From a deployment perspective, three practical implications follow: (1) bounded automation: reminder scheduling (and any other task execution) should be gated by confirmation and explicit ambiguity handling; (2) uncertainty-aware interaction: low-confidence parsing, ambiguous times, or weak retrieval alignment should trigger clarification rather than completion; and (3) auditability: stored notes, retrieved evidence, and scheduled reminders should be traceable (e.g., via interaction IDs and log records) to support accountability, incident review, and governance. These implications follow directly from the observed reminder trade-offs and the dependence of downstream performance on upstream correctness.

\subsection{Limitations}

This study has a few limitations that should be considered when interpreting the results. First, the care-home trials were conducted in two sites only and involved a limited number of supervised interactions. While these trials provided valuable empirical data, they do not capture the full range of variability present across different care homes, staff roles, or working practices. Second, the system was evaluated as a prototype decision-support tool under supervision. Outputs generated by the system were not used to influence real care delivery, and the evaluation did not assess long-term use, learning effects, or sustained adoption over time. Third, synthetic records were used to populate the database for retrieval evaluation and stress-testing. Although these records were structured to resemble typical care documentation, they may not fully reflect the complexity, variability, or inconsistencies present in real-world care records. Fourth, retrieval performance was evaluated using a defined set of spoken queries and logged outcomes. The reported retrieval performance reflects behaviour under the evaluated conditions and query formulations and should not be interpreted as a guarantee of performance in all usage scenarios. Finally, this study did not assess clinical outcomes, staff workload impact, or user satisfaction in a systematic way. Such assessments would require longer-term, in-situ studies and were outside the scope of the present evaluation.

Additional limitations that should be stated explicitly to match the safety-focused framing:
\begin{itemize}
\item Latency and availability were not measured end-to-end, so real-time responsiveness and resilience to downtime remain unvalidated under deployment conditions.
\item No systematic stratification was performed by accent group, background noise level, or speech impairment, limiting claims about inclusivity and robustness beyond the evaluated sample.
\end{itemize}

\section{Conclusion and Future Work}

This paper presented an assurance-driven evaluation of a multi-agent, voice-enabled smart speaker designed to support care documentation, information retrieval, and reminder scheduling in care home settings. Using supervised trials conducted in two UK care homes and controlled testing, the study examined system behaviour across the full processing pipeline, from speech recognition to structured storage, retrieval, and task scheduling. Rather than focusing solely on component-level performance, the evaluation framework emphasised end-to-end behaviour, uncertainty management, and evidence-based reasoning about system risk. The results indicate that high reliability can be achieved for structured extraction tasks (notably resident identifier and care category matching) under the evaluated conditions, supporting accurate database insertion and subsequent access to recorded information. In contrast, reminder handling remains the primary risk concentration, reflecting the difficulty of converting informal spoken instructions into precise, schedulable actions. These findings reinforce the need for bounded automation, particularly confirmation and clarification prompts, to prevent uncertain reminder requests from being executed silently. Overall, the study demonstrates how quantitative metrics with uncertainty bounds and an assurance-style argument can be combined to provide a transparent and defensible evaluation of voice-enabled systems intended for use in safety-critical care environments. This work contributes a structured approach for evaluating conversational AI in care contexts and supports the responsible development of voice-enabled technologies as decision-support tools within health and social care. Future work should extend validation through larger and longer in-situ trials, include stratified analysis across accents/noise/speech profiles, and strengthen retrieval and scheduling evaluation with ranking metrics and field-level correctness checks.

Future work should extend this evaluation through larger-scale, longer-duration trials across a broader range of care settings. Additional evaluation could examine how system performance evolves over time, how staff adapt their interaction patterns, and how voice-enabled systems integrate with existing care documentation practices. Further work is also needed to examine governance, oversight, and accountability considerations associated with deploying voice-enabled smart speaker systems in care environments, particularly as systems move beyond supervised trials toward routine use. Methodologically, future evaluation should (i) measure field-level scheduling correctness (what/when/who), (ii) add retrieval ranking metrics (Recall@k/MRR) and tail-risk reporting, (iii) stratify performance by accent/noise/speech profiles, and (iv) validate safe-deferral policies by reporting clarification/confirmation rates and their impact on staff workload.

\section{Author's Contribution}
Zeinab Dehghani, Rameez Raja Kureshi, Koorosh Aslansefat, Faezeh Abedi, Dhavalkumar Thakker, and Lisa Greaves contributed to the study's conception and design; data collection, analysis, system conceptualisation, design, development, validation, and writing. Bhupesh Kumar Mishra, Baseer Ahmad and Tanaya Maslekar contributed to the study's system validation and writing.

\section{Acknowledgement}

The research and development work in this project was funded by Innovate UK and Connexin, Hull, as part of a Knowledge Transfer Partnership (KTP) between the University of Hull, UK and Connexin.

\bibliographystyle{plain}

\begin{thebibliography}{10}

\bibitem{adedeji2024sound}
Ayo Adedeji, Sarita Joshi, and Brendan Doohan.
\newblock The sound of healthcare: Improving medical transcription asr accuracy with large language models.
\newblock {\em arXiv preprint arXiv:2402.07658}, 2024.

\bibitem{amershi2019software}
Saleema Amershi, Andrew Begel, Christian Bird, Robert DeLine, Harald Gall, Ece Kamar, Nachiappan Nagappan, Besmira Nushi, and Thomas Zimmermann.
\newblock Software engineering for machine learning: A case study.
\newblock In {\em 2019 IEEE/ACM 41st International Conference on Software Engineering: Software Engineering in Practice (ICSE-SEIP)}, pages 291--300. IEEE, 2019.

\bibitem{astell2024like}
Arlene Astell and David Clayton.
\newblock “like another human being in the room”: a community case study of smart speakers to reduce loneliness in the oldest-old.
\newblock {\em Frontiers in Psychology}, 15:1320555, 2024.

\bibitem{aston2025smartcarehome}
{Aston University}.
\newblock First ai-powered smart care home system to improve quality of residential care.
\newblock University news article, 2025.
\newblock Accessed: 2026-03-06.

\bibitem{ballati2018hey}
Fabio Ballati, Fulvio Corno, Luigi De~Russis, et~al.
\newblock " hey siri, do you understand me?": Virtual assistants and dysarthria.
\newblock In {\em Intelligent Environments (Workshops)}, pages 557--566, 2018.

\bibitem{bates2021potential}
David~W Bates, David Levine, Ania Syrowatka, Masha Kuznetsova, Kelly Jean~Thomas Craig, Angela Rui, Gretchen~Purcell Jackson, and Kyu Rhee.
\newblock The potential of artificial intelligence to improve patient safety: a scoping review.
\newblock {\em NPJ digital medicine}, 4(1):54, 2021.

\bibitem{belisle2024stakeholder}
Jean-Christophe B{\'e}lisle-Pipon, Maria Powell, Renee English, Marie-Fran{\c{c}}oise Malo, Vardit Ravitsky, Bridge2AI-Voice Consortium, and Yael Bensoussan.
\newblock Stakeholder perspectives on ethical and trustworthy voice ai in health care.
\newblock {\em Digital Health}, 10:20552076241260407, 2024.

\bibitem{bentley2018understanding}
Frank Bentley, Chris Luvogt, Max Silverman, Rushani Wirasinghe, Brooke White, and Danielle Lottridge.
\newblock Understanding the long-term use of smart speaker assistants.
\newblock {\em Proceedings of the ACM on Interactive, Mobile, Wearable and Ubiquitous Technologies}, 2(3):1--24, 2018.

\bibitem{bishop2006pattern}
Christopher~M Bishop and Nasser~M Nasrabadi.
\newblock {\em Pattern recognition and machine learning}, volume~4.
\newblock Springer, 2006.

\bibitem{bjerkan2021patient}
Jorunn Bjerkan, Victor Valderaune, and Rose~Mari Olsen.
\newblock Patient safety through nursing documentation: Barriers identified by healthcare professionals and students.
\newblock {\em Frontiers in Computer Science}, 3:624555, 2021.

\bibitem{bonilla2020older}
Karen Bonilla and Aqueasha Martin-Hammond.
\newblock Older adults’ perceptions of intelligent voice assistant privacy, transparency, and online privacy guidelines.
\newblock In {\em Sixteenth symposium on usable privacy and security (SOUPS 2020)}, 2020.

\bibitem{burton2020mind}
Simon Burton, Ibrahim Habli, Tom Lawton, John McDermid, Phillip Morgan, and Zoe Porter.
\newblock Mind the gaps: Assuring the safety of autonomous systems from an engineering, ethical, and legal perspective.
\newblock {\em Artificial Intelligence}, 279:103201, 2020.

\bibitem{calinescu2017engineering}
Radu Calinescu, Danny Weyns, Simos Gerasimou, Muhammad~Usman Iftikhar, Ibrahim Habli, and Tim Kelly.
\newblock Engineering trustworthy self-adaptive software with dynamic assurance cases.
\newblock {\em IEEE Transactions on Software Engineering}, 44(11):1039--1069, 2017.

\bibitem{carrick2025speech}
Jonathan~E Carrick, Nina Dethlefs, Lisa Greaves, Venkata~MV Gunturi, Rameez~Raja Kureshi, and Yongqiang Cheng.
\newblock Speech-controlled smart speaker for accurate, real-time health and care record management.
\newblock In {\em Proceedings of the 15th International Workshop on Spoken Dialogue Systems Technology}, pages 238--244, 2025.

\bibitem{chen2023older}
Chen Chen, Ella~T Lifset, Yichen Han, Arkajyoti Roy, Michael Hogarth, Alison~A Moore, Emilia Farcas, and Nadir Weibel.
\newblock How do older adults set up voice assistants? lessons learned from a deployment experience for older adults to set up standalone voice assistants.
\newblock In {\em Companion Publication of the 2023 ACM Designing Interactive Systems Conference}, pages 164--168, 2023.

\bibitem{edwards2021use}
Katie~J Edwards, Ray~B Jones, Deborah Shenton, Toni Page, Inocencio Maramba, Alison Warren, Fiona Fraser, Tanja Kri{\v{z}}aj, Tristan Coombe, Hazel Cowls, et~al.
\newblock The use of smart speakers in care home residents: implementation study.
\newblock {\em Journal of medical internet research}, 23(12):e26767, 2021.

\bibitem{ehrlich2025forestgpt}
Florian Ehrlich-Sommer, Benno Eberhard, and Andreas Holzinger.
\newblock Forestgpt and beyond: A trustworthy domain-specific large language model paving the way to forestry 5.0.
\newblock {\em Electronics}, 14(18):3583, 2025.

\bibitem{franco2023exploratory}
Jessica Franco, Yauheni Solad, Arjun~K Venkatesh, Reinier Van~Tonder, Alexander Solod, Tomek Stachowiak, Allen~L Hsiao, and Rohit~B Sangal.
\newblock Exploratory descriptive analysis of smart speaker utilization in the emergency department during the covid-19 pandemic.
\newblock {\em The Journal of Emergency Medicine}, 64(4):506--512, 2023.

\bibitem{green2003automatic}
Phil~D Green, James Carmichael, Athanassios Hatzis, Pam Enderby, Mark~S Hawley, and Mark Parker.
\newblock Automatic speech recognition with sparse training data for dysarthric speakers.
\newblock In {\em Interspeech}, pages 1189--1192, 2003.

\bibitem{hawkins2021guidance}
Richard Hawkins, Colin Paterson, Chiara Picardi, Yan Jia, Radu Calinescu, and Ibrahim Habli.
\newblock Guidance on the assurance of machine learning in autonomous systems (amlas).
\newblock {\em arXiv preprint arXiv:2102.01564}, 2021.

\bibitem{hui2025enhancing}
Macarious Hui, Jinda Zhang, and Aanchan Mohan.
\newblock Enhancing aac software for dysarthric speakers in e-health settings: an evaluation using torgo.
\newblock In {\em ICC 2025-IEEE International Conference on Communications}, pages 3673--3679. IEEE, 2025.

\bibitem{jurafsky2000speech}
Daniel Jurafsky and James~H Martin.
\newblock Speech and language processing. an introduction to nlp, computational linguistics, and speech recognition.
\newblock {\em Computational Linguistics, and Speech Recognition, Prentice Hall}, 2, 2000.

\bibitem{kelly2004goal}
Tim Kelly and Rob Weaver.
\newblock The goal structuring notation--a safety argument notation.
\newblock In {\em Proceedings of the dependable systems and networks 2004 workshop on assurance cases}, volume~6. Citeseer Princeton, NJ, 2004.

\bibitem{kim2021exploring}
Sunyoung Kim and Abhishek Choudhury.
\newblock Exploring older adults’ perception and use of smart speaker-based voice assistants: A longitudinal study.
\newblock {\em Computers in Human Behavior}, 124:106914, 2021.

\bibitem{koenecke2020racial}
Allison Koenecke, Andrew Nam, Emily Lake, Joe Nudell, Minnie Quartey, Zion Mengesha, Connor Toups, John~R Rickford, Dan Jurafsky, and Sharad Goel.
\newblock Racial disparities in automated speech recognition.
\newblock {\em Proceedings of the national academy of sciences}, 117(14):7684--7689, 2020.

\bibitem{kusner2015word}
Matt Kusner, Yu~Sun, Nicholas Kolkin, and Kilian Weinberger.
\newblock From word embeddings to document distances.
\newblock In {\em International conference on machine learning}, pages 957--966. PMLR, 2015.

\bibitem{laranjo2018conversational}
Liliana Laranjo, Adam~G Dunn, Huong~Ly Tong, Ahmet~Baki Kocaballi, Jessica Chen, Rabia Bashir, Didi Surian, Blanca Gallego, Farah Magrabi, Annie~YS Lau, et~al.
\newblock Conversational agents in healthcare: a systematic review.
\newblock {\em Journal of the American Medical Informatics Association}, 25(9):1248--1258, 2018.

\bibitem{lau2018alexa}
Josephine Lau, Benjamin Zimmerman, and Florian Schaub.
\newblock Alexa, are you listening? privacy perceptions, concerns and privacy-seeking behaviors with smart speakers.
\newblock {\em Proceedings of the ACM on human-computer interaction}, 2(CSCW):1--31, 2018.

\bibitem{lewis2020retrieval}
Patrick Lewis, Ethan Perez, Aleksandra Piktus, Fabio Petroni, Vladimir Karpukhin, Naman Goyal, Heinrich K{\"u}ttler, Mike Lewis, Wen-tau Yih, Tim Rockt{\"a}schel, et~al.
\newblock Retrieval-augmented generation for knowledge-intensive nlp tasks.
\newblock {\em Advances in neural information processing systems}, 33:9459--9474, 2020.

\bibitem{lima2025promoting}
Maria~R Lima, Amy O'Connell, Feiyang Zhou, Alethea Nagahara, Avni Hulyalkar, Anura Deshpande, Jesse Thomason, Ravi Vaidyanathan, and Maja Matari{\'c}.
\newblock Promoting cognitive health in elder care with large language model-powered socially assistive robots.
\newblock In {\em Proceedings of the 2025 CHI Conference on Human Factors in Computing Systems}, pages 1--22, 2025.

\bibitem{merkel2025}
Sebastian Merkel and Sabrina Schorr.
\newblock Identification of use cases, target groups, and motivations around adopting smart speakers for health care and social care settings: Scoping review.
\newblock {\em JMIR AI}, 4(1):e55673, 2025.

\bibitem{merkel2025identification}
Sebastian Merkel, Sabrina Schorr, et~al.
\newblock Identification of use cases, target groups, and motivations around adopting smart speakers for health care and social care settings: Scoping review.
\newblock {\em JMIR AI}, 4(1):e55673, 2025.

\bibitem{mittelstadt2019principles}
Brent Mittelstadt.
\newblock Principles alone cannot guarantee ethical ai.
\newblock {\em Nature machine intelligence}, 1(11):501--507, 2019.

\bibitem{nimrod2022technology}
Galit Nimrod and Yael Edan.
\newblock Technology domestication in later life.
\newblock {\em International Journal of Human--Computer Interaction}, 38(4):339--350, 2022.

\bibitem{oakland2024voicepilot}
{Oakland Care}.
\newblock Oakland care announces state-of-the-art voice tech trial.
\newblock Company news release, 2024.
\newblock Accessed: 2026-03-06.

\bibitem{okonji2024applications}
Onyekachukwu~R Okonji, Kamol Yunusov, and Bonnie Gordon.
\newblock Applications of generative ai in healthcare: algorithmic, ethical, legal and societal considerations.
\newblock {\em arXiv preprint arXiv:2406.10632}, 2024.

\bibitem{pacyna2025patient}
JE~Pacyna, MD~Anzabi, AM~Stroud, JL~Wise, and RR~Sharp.
\newblock Patient concerns about ai-based voice analysis in healthcare.
\newblock {\em BMC Digital Health}, 3(1):79, 2025.

\bibitem{powers2020evaluation}
David~MW Powers.
\newblock Evaluation: from precision, recall and f-measure to roc, informedness, markedness and correlation.
\newblock {\em arXiv preprint arXiv:2010.16061}, 2020.

\bibitem{pradhan2018accessibility}
Alisha Pradhan, Kanika Mehta, and Leah Findlater.
\newblock " accessibility came by accident" use of voice-controlled intelligent personal assistants by people with disabilities.
\newblock In {\em Proceedings of the 2018 CHI Conference on human factors in computing systems}, pages 1--13, 2018.

\bibitem{quinn2024assessing}
Kelly Quinn, Sarah Leiser~Ransom, Carrie O'Connell, Naoko Muramatsu, David~X Marquez, and Jessie Chin.
\newblock Assessing the feasibility and acceptability of smart speakers in behavioral intervention research with older adults: Mixed methods study.
\newblock {\em Journal of medical Internet research}, 26:e54800, 2024.

\bibitem{raghu2020survey}
Maithra Raghu and Eric Schmidt.
\newblock A survey of deep learning for scientific discovery.
\newblock {\em arXiv preprint arXiv:2003.11755}, 2020.

\bibitem{saripalle2024command}
Rishi Saripalle and Ravi Patel.
\newblock From command to care: A scoping review on utilization of smart speakers by patients and providers.
\newblock {\em Mayo Clinic Proceedings: Digital Health}, 2(2):207--220, 2024.

\bibitem{saripalle2024}
Rishi Saripalle and Ravi Patel.
\newblock From command to care: A scoping review on utilization of smart speakers by patients and providers.
\newblock {\em Mayo Clinic Proceedings: Digital Health}, 2(2):134--147, 2024.

\bibitem{sarvari2025challenges}
Peter Sarvari, Zaid Al-Fagih, Alexander Abou-Chedid, Paul Jewell, Rosie Taylor, and Arouba Imtiaz.
\newblock Challenges and solutions in applying large language models to guideline-based management planning and automated medical coding in health care: Algorithm development and validation.
\newblock {\em JMIR Biomedical Engineering}, 10(1):e66691, 2025.

\bibitem{schutze2008introduction}
Hinrich Sch{\"u}tze, Christopher~D Manning, and Prabhakar Raghavan.
\newblock {\em Introduction to information retrieval}, volume~39.
\newblock Cambridge University Press Cambridge, 2008.

\bibitem{sendak2020path}
Mark~P Sendak, Joshua D’Arcy, Sehj Kashyap, Michael Gao, Marshall Nichols, Kristin Corey, William Ratliff, and Suresh Balu.
\newblock A path for translation of machine learning products into healthcare delivery.
\newblock {\em EMJ Innov}, 10:19--00172, 2020.

\bibitem{smith2023smart}
Elizabeth Smith, Petroc Sumner, Craig Hedge, and Georgina Powell.
\newblock Smart-speaker technology and intellectual disabilities: agency and wellbeing.
\newblock {\em Disability and Rehabilitation: Assistive Technology}, 18(4):432--442, 2023.

\bibitem{topol2019deep}
Eric Topol.
\newblock {\em Deep medicine: how artificial intelligence can make healthcare human again}.
\newblock Hachette UK, 2019.

\bibitem{treder2024introduction}
Matthias~S Treder, Sojin Lee, and Kamen~A Tsvetanov.
\newblock Introduction to large language models (llms) for dementia care and research.
\newblock {\em Frontiers in dementia}, 3:1385303, 2024.

\bibitem{van2002distributed}
Maarten Van~Steen.
\newblock Distributed systems principles and paradigms.
\newblock {\em Network}, 2(28):1, 2002.

\bibitem{wilson1927probable}
Edwin~B Wilson.
\newblock Probable inference, the law of succession, and statistical inference.
\newblock {\em Journal of the American Statistical Association}, 22(158):209--212, 1927.

\bibitem{yang2024talk2care}
Ziqi Yang, Xuhai Xu, Bingsheng Yao, Ethan Rogers, Shao Zhang, Stephen Intille, Nawar Shara, Guodong~Gordon Gao, and Dakuo Wang.
\newblock Talk2care: An llm-based voice assistant for communication between healthcare providers and older adults.
\newblock {\em Proceedings of the ACM on Interactive, Mobile, Wearable and Ubiquitous Technologies}, 8(2):1--35, 2024.

\end{thebibliography}

\end{document}